\documentclass[journal]{IEEEtran}

\usepackage{xcolor}
\usepackage{soul}  
\newcommand{\red}[1]{{\color{red}#1}}
\newcommand{\todo}[1]{{\color{red}#1}}
\newcommand{\TODO}[1]{\textbf{\color{red}[TODO: #1]}}

\newcommand{\karen}[1]{\textcolor{red}{[Karen: #1]}}
\newcommand{\ze}[1]{\textcolor{green}{[Yanjie: #1]}}
\newcommand{\ken}[1]{\textcolor{blue}{[Ken: #1]}}
\newcommand{\monroe}[1]{\textcolor{magenta}{[Monroe: #1]}}

\newif\ifhidecomments
\ifhidecomments
  \renewcommand{\TODO}[1]{}
  \renewcommand{\todo}[1]{}
  \renewcommand{\red}[1]{}
  \renewcommand{\karen}[1]{}
  \renewcommand{\ze}[1]{}
  \renewcommand{\ken}[1]{}
  \renewcommand{\monroe}[1]{}
\fi

\usepackage{amsmath}
\usepackage{amssymb}
\usepackage{graphicx}
\usepackage{comment}
\usepackage{float}
\usepackage{booktabs}
\usepackage{cite}
\usepackage[caption=false,font=footnotesize]{subfig}
\usepackage[breaklinks,colorlinks,citecolor=blue,urlcolor=blue,linkcolor=blue]{hyperref}
\usepackage[capitalize]{cleveref}
\usepackage{stfloats}

\title{Learning Humanoid Navigation from Human Data}

\author{Weizhuo~Wang,
        Yanjie~Ze,
        C.~Karen~Liu,
        and~Monroe~Kennedy~III,
\thanks{W. Wang, Y. Ze, C. K. Liu, and M. Kennedy III are with Stanford University, Stanford, CA 94305 USA (e-mail: weizhuo2@stanford.edu; yanjieze@stanford.edu; ckliu38@stanford.edu; monroek@stanford.edu).}}%

\begin{document}
\maketitle



\begin{abstract}
    We present EgoNav, a system that enables a humanoid robot to traverse diverse, unseen environments by learning entirely from 5 hours of human walking data, with no robot data or finetuning. A diffusion model predicts distributions of plausible future trajectories conditioned on past trajectory, a 360\textdegree{} visual memory fusing color, depth, and semantics, and video features from a frozen DINOv3 backbone that capture appearance cues invisible to depth sensors. A hybrid sampling scheme achieves real-time inference in 10 denoising steps, and a receding-horizon controller selects paths from the predicted distribution. We validate EgoNav through offline evaluations, where it outperforms baselines in collision avoidance and multi-modal coverage, and through zero-shot deployment on a Unitree G1 humanoid across unseen indoor and outdoor environments. Behaviors such as waiting for doors to open, navigating around crowds, and avoiding glass walls emerge naturally from the learned prior. We will release the dataset and trained models. Our website: \url{https://egonav.weizhuowang.com/}
\end{abstract}

\section{Introduction}
\label{sec:intro}

\begin{figure*}[!b]
    \centering
    \includegraphics[width=\textwidth]{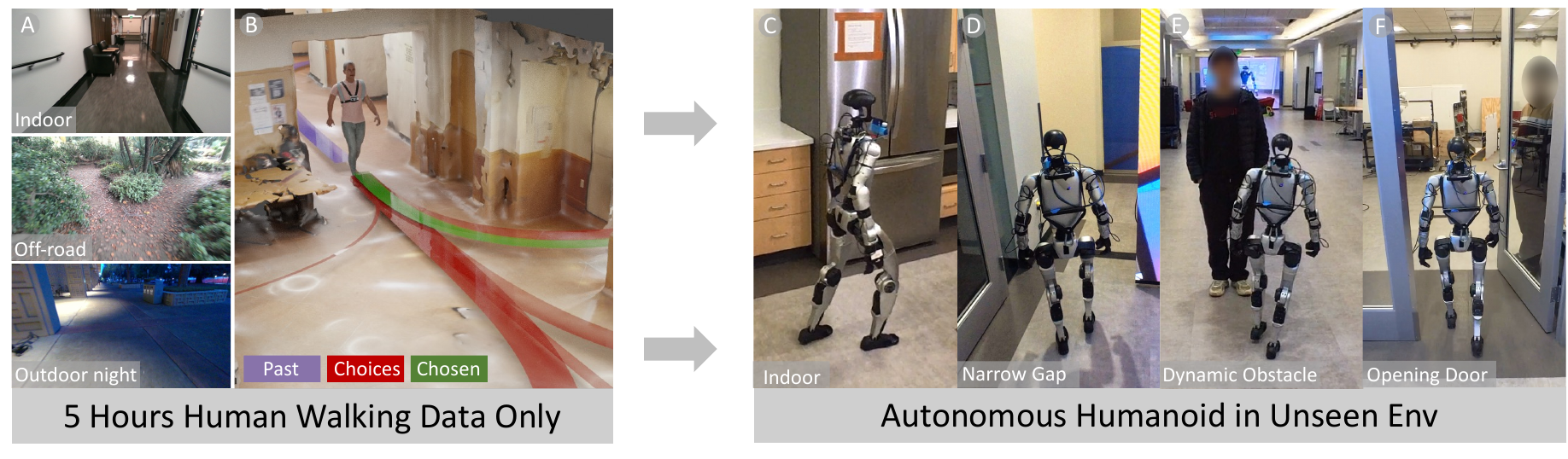}
    \vspace{-6mm}

    \caption{\textbf{EgoNav} learns a navigation prior from human walking data: given past trajectory (purple) and a 360\textdegree{} visual context of the scene, a diffusion model generates a distribution of plausible future paths (red ribbons), with ribbon width indicating likelihood. The learned prior transfers directly to a Unitree G1 humanoid with zero robot data.}
    \label{fig:outline}

\end{figure*}

How can a humanoid robot learn to navigate diverse, unseen environments without explicitly building a map? Previous work on learning-based robot navigation collected data directly on robot platforms, which is costly and difficult to scale. Cross-embodiment approaches such as NoMaD \cite{sridhar_nomad_2023} train navigation policies on hundreds of hours of robot driving data and demonstrate transfer across wheeled and legged platforms, but each new domain still requires in-domain training data, and large-scale humanoid navigation datasets do not yet exist.

Vision-language models such as NaVILA \cite{cheng_navila_2025} sidestep robot data by using a VLM to translate language instructions into locomotion commands. However, they require step-by-step human instructions, emit discrete unimodal commands without spatial grounding, and are too slow for closed-loop control.

In manipulation, a promising alternative has emerged: learning from human demonstrations. Systems such as UMI \cite{chi_universal_2024} and DexCap \cite{wang_dexcap_2024} train policies entirely on human data and transfer them to robots without any robot-specific data collection. Navigation is a natural next frontier for this paradigm. Human walking data is cheap and scalable: a single person with body-mounted cameras can gather hours of diverse navigation data without any robot hardware. Human walking data also encodes rich commonsense about navigation: which paths are physically plausible, how to avoid obstacles, and where alternative routes exist. Despite this promise, recent attempts to learn navigation from human data \cite{chen_hand-eye_2025} still struggle to work without robot data and show limited generalization. 

What prevents this paradigm from working in the real world? We identify some key challenges that must be addressed:

\begin{itemize}
    \item \textbf{Insufficient scene coverage:} Scene coverage is twofold: one is the field of view (FOV) of the camera, the other is the richness of the scene description. Most trajectory prediction approaches assume ground truth scenes \cite{cui_multimodal_2019, varadarajan_multipath_2022,rempe_trace_2023} or predict from past trajectories alone \cite{wiest_probabilistic_2012, gu_stochastic_2022, lv_learning_2024}. The egocentric perspective is more practical, but a stereo depth camera covers only $\sim$90\textdegree{} FOV, is blind to transparent or reflective surfaces such as glass walls, and carries no semantic understanding of the scene.
    \item \textbf{Unimodal predictions:} Human motion is inherently multi-modal: at any decision point, multiple future trajectories are plausible. Yet most methods produce single-trajectory estimates \cite{helbing_social_1995, wang_trajectory_2023, pan_lookout_2025, qiu_egocognav_2025}, providing incomplete information for downstream path selection. A useful navigation prior must capture the diverse distribution of plausible futures, not a single guess.
    \item \textbf{Interface representation:} Some prior work uses end-to-end methods that couple terrain traversal and locomotion, making it hard to generalize in the real world. Others predict hand and eye trajectories in the robot frame, feeding to a decoupled locomotion policy, but despite careful engineering to bridge the embodiment gap, find that human data alone is insufficient and must be supplemented with robot demonstrations. A successful human-to-robot transfer requires an embodiment-agnostic interface between navigation and locomotion, one that conveys not only \emph{where} to go but also \emph{what the path looks like} semantically, so that downstream controllers can adapt their behavior to the terrain.
\end{itemize}

The key is to learn an embodiment-agnostic \emph{navigation prior}: a distribution over plausible future trajectories in world coordinates conditioned on the surrounding scene. In this paper, we present EgoNav, a system that addresses these challenges and learns a generalizable navigation prior from human walking data for humanoid deployment (\cref{fig:outline}):

\begin{itemize}
    \item We construct a 360\textdegree{} panoramic \textbf{visual memory} (VM) fusing color, depth, and semantic channels from a rolling buffer of egocentric observations, augmented with video features from a frozen DINOv3 backbone \cite{simeoni_dinov3_2025} that capture appearance-level cues invisible to depth sensors, such as glass walls and dynamic agents.
    \item We train a \textbf{conditional diffusion model} on the VM and video features that generates diverse future trajectory samples, naturally capturing the multi-modal distribution of plausible paths. To overcome the latency barrier of iterative diffusion sampling for real-time deployment \cite{rempe_trace_2023, jia_learning_2025}, we introduce a hybrid DDIM--DDPM sampling scheme that achieves near-full quality in only 10 steps.
    \item We build a complete \textbf{pipeline from human data to humanoid deployment} on a Unitree G1: a receding-horizon controller selects from the predicted distribution with latency compensation and mode-consistency.
\end{itemize}

We emphasize that EgoNav learns a \emph{navigation prior}: a scene-informed distribution of plausible paths from which a downstream controller selects and executes. High-level reasoning (VLMs) and low-level locomotion are advancing rapidly, but the middle layer, knowing \emph{where one can walk}, remains missing. A robust, generalizable navigation prior fills this gap, onto which goals or task planning can be layered. The resulting navigation prior, trained entirely on 5 hours of human walking data with no robot data or finetuning, transfers directly to autonomous humanoid navigation in unseen environments. We validate EgoNav in offline evaluations, demonstrating superior collision avoidance and multi-modal trajectory coverage over baselines. Through real-world deployment on the G1 humanoid in previously unseen environments, we show that behaviors such as waiting for doors to open, navigating around crowds, and avoiding glass walls emerge from the learned prior without explicit programming. The dataset and trained models will be publicly released.
\section{Related Work}
\label{sec:relatedwork}
Our work learns humanoid navigation from human walking data, drawing on advances in scene understanding, multi-modal generative modeling, and human-to-robot transfer for deployment. We review related work across these areas.
\begin{figure}[tb]
  \centering
  \includegraphics[width=\linewidth]{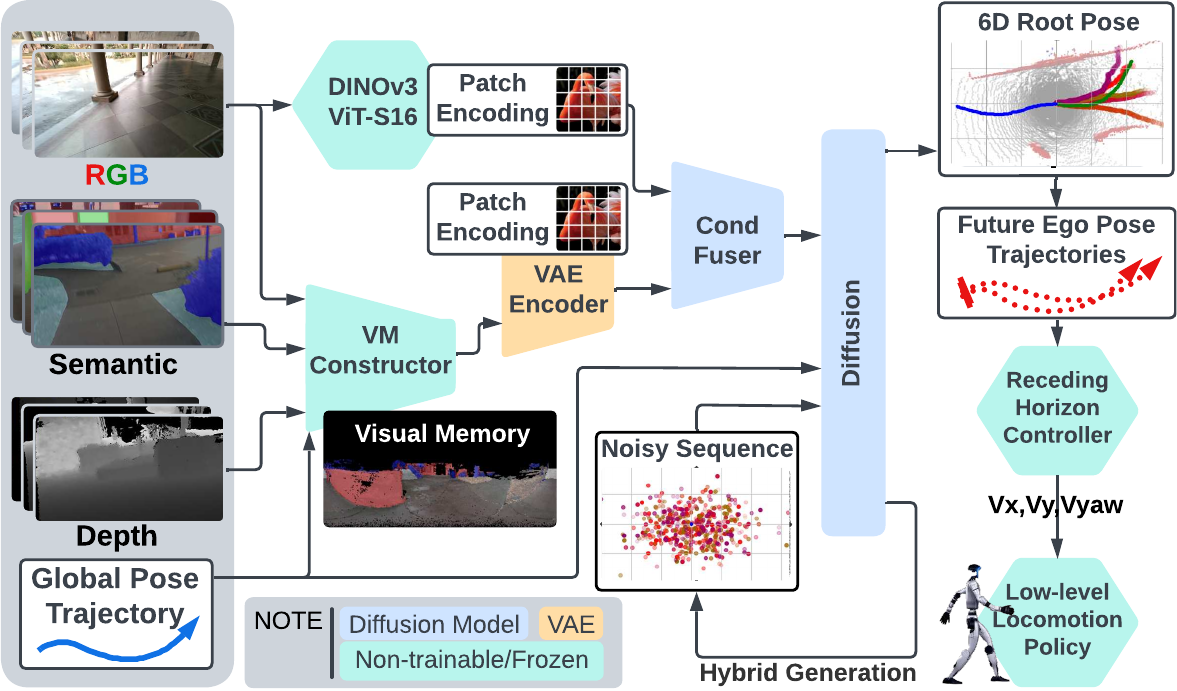}
  \caption{\textbf{Overview of the proposed method:} A rolling buffer of 32 segmented RGB frames and cleaned depth frames are combined to construct a single visual memory (VM). The VM is encoded into a 64-dimensional embedding, fused with DINOv3 video features, and concatenated with 6D pose as input to the diffusion model. Then the future trajectories are denoised. All input and output of the prediction module are in the egocentric frame.}
  \label{fig:sys_overview}
\vspace{-2mm}
\end{figure}

\textbf{Trajectory prediction for autonomous systems.}
In autonomous driving, trajectory prediction is essential for safe planning \cite{rempe_trace_2023}. Early recurrent approaches \cite{tang_multiple_2019} have given way to transformer-based models \cite{mercat_multi-head_2019, vaswani_attention_2023} that better handle diverse sensor streams and long-range dependencies. AV methods commonly represent the environment in bird's-eye-view (BEV) or occupancy grids \cite{strohbeck_multiple_2020, schreiber_long-term_2019, chai_multipath_2019}, which simplify prediction under structured traffic rules but struggle in unstructured settings \cite{rempe_generating_2022}.
For pedestrian trajectory prediction, classical approaches such as Social Force \cite{helbing_social_1995}, Social GAN \cite{gupta_social_2018}, and Social LSTM \cite{alahi_social_2016} model interactions between agents but largely ignore environmental context, predicting solely from past trajectories \cite{wiest_probabilistic_2012, gu_stochastic_2022, lv_learning_2024}. Prominent datasets such as UCY/ETH \cite{lerner_crowds_2007} and the Stanford Drone Dataset \cite{robicquet_learning_2016} provide only BEV observations, reducing the problem to two dimensions. While effective offline, the lack of scene awareness and reliance on BEV limit real-world deployment.

\textbf{Egocentric scene representations.}
Recent work has shifted toward richer scene representations for trajectory prediction. While AV methods \cite{cui_multimodal_2019, varadarajan_multipath_2022, chen_scept_2022, sun_large_2024} rely on synthesized BEV, walking scenarios demand more flexible egocentric representations. Early approaches use first-person RGB(D) directly \cite{singh_krishnacam_2016, park_egocentric_2016}, but single frames provide limited spatial coverage and lose context outside the camera frustum. Wang et al.\ \cite{wang_trajectory_2023} address this with a depth panorama built from a rolling buffer of egocentric observations, extending the effective field of view to 360\textdegree{}, though without color or semantic information. More recently, vision foundation models such as DINOv3 \cite{simeoni_dinov3_2025} have emerged as powerful feature extractors, capturing appearance and semantic cues that complement geometric depth. Our visual memory builds on the panoramic paradigm by fusing color, depth, and semantic channels into a single compact representation, augmented with DINOv3 features.

\textbf{Diffusion models for trajectory prediction.}
Denoising diffusion probabilistic models (DDPM) \cite{ho_denoising_2020} and their accelerated variant DDIM \cite{song_denoising_2022} have become a standard framework for multi-modal generation. In the trajectory domain, TRACE \cite{rempe_trace_2023} applies guided diffusion to generate diverse future human motions in 3D scenes, demonstrating clear advantages over deterministic predictions in capturing multi-modal distributions, but operates offline without real-time constraints. Jia et al.\ \cite{jia_learning_2025} use DDPM to predict egocentric upper-body motion including head pose and gaze, though their focus is on visuomotor coordination rather than locomotion trajectories. A common limitation of diffusion-based prediction methods is inference speed: the iterative denoising process is typically too slow for real-time robotic applications. Our method addresses this through a hybrid DDIM--DDPM sampling scheme that achieves real-time performance with minimal degradation in as few as 10 steps, while using classifier-free guidance to condition generation on scene context and video features.

\textbf{Egocentric trajectory prediction.}
A growing body of work addresses motion prediction from the egocentric perspective. LookOut \cite{pan_lookout_2025} lifts egocentric DINOv2 features into a 3D voxel grid, collapses it to a bird's-eye view, and regresses future 6D head poses, but produces only deterministic single-trajectory estimates. EgoCogNav \cite{qiu_egocognav_2025} jointly predicts egocentric body-frame trajectories and perceived path uncertainty using DINOv2 features fused with gaze and motion cues, introducing a cognition-aware dimension; however, it also produces single-trajectory estimates and does not deploy on a robot. HEAD \cite{chen_hand-eye_2025} learns deterministic humanoid goal-conditioned navigation on a Unitree G1 robot by predicting hand and eye trajectories. Despite careful engineering (image undistortion, homography alignment, temporal subsampling) to address the embodiment gap, the authors find that human data alone is insufficient and must be supplemented with robot demonstrations, showing limited generalization within room scale. The primary issue is that hand and eye trajectories remain deeply coupled to the robot embodiment.

\textbf{Robot navigation and human-to-robot transfer.}
In manipulation, recent works such as UMI \cite{chi_universal_2024} and DexCap \cite{wang_dexcap_2024} have demonstrated that policies trained on human demonstration data can transfer directly to robots without robot-specific data collection. In navigation, NoMaD \cite{sridhar_nomad_2023} trains a diffusion-based policy on over 100 hours of pure robot driving data and demonstrates cross-platform transfer, but outputs only short-horizon 2D waypoints from a single forward-facing camera with no semantic scene understanding. More fundamentally, it still requires extensive robot data collection. The paradigm of learning navigation from human data remains largely unexplored. VP-Nav \cite{fu_coupling_2022} develops a point-goal navigation system for quadrupeds that couples vision with proprioceptive feedback, enabling the robot to detect obstacles invisible to depth sensors such as glass walls. Its classical planning pipeline (occupancy map + FMM) is environment-agnostic but purely geometric, with no semantic scene understanding. ANYmal Parkour \cite{hoeller_anymal_2023} and Locomotion Beyond Feet \cite{yang_locomotion_2026} demonstrate agile legged locomotion through libraries of terrain-specific skills (jumping, climbing, crouching), but only generalize to novel arrangements of known obstacle categories. Concurrent advances in general humanoid locomotion \cite{radosavovic_humanoid_2024, hu_robot_2025} have made real-world deployment increasingly feasible, but these controllers solve \emph{how} to walk without addressing \emph{where} to walk. Our system fills this gap: trained entirely on human walking data with no robot data or finetuning, it provides a generalizable navigation prior that transfers directly to a humanoid robot.

Our work differs from the above in a fundamental way: we require \emph{no robot data and no finetuning}. While NoMaD depends on large-scale purely robot-collected datasets and HEAD must supplement human data with robot demonstrations, our model, trained entirely on human walking demonstrations, predicts embodiment-agnostic 6D trajectory distributions and transfers directly to a Unitree G1 humanoid. Beyond this data paradigm shift, we provide (1) a 360\textdegree{} panoramic visual memory augmented with DINOv3 video features for richer scene understanding than forward-facing cameras alone, (2) long-horizon multi-modal trajectory distributions (5\,s, 20Hz, 6DoF) rather than short-horizon action predictions, and (3) real-world humanoid deployment validated across multiple unseen environments.
\section{Method}

\label{sec:method}

\begin{figure}[tb]
  \centering
  \includegraphics[width=\columnwidth]{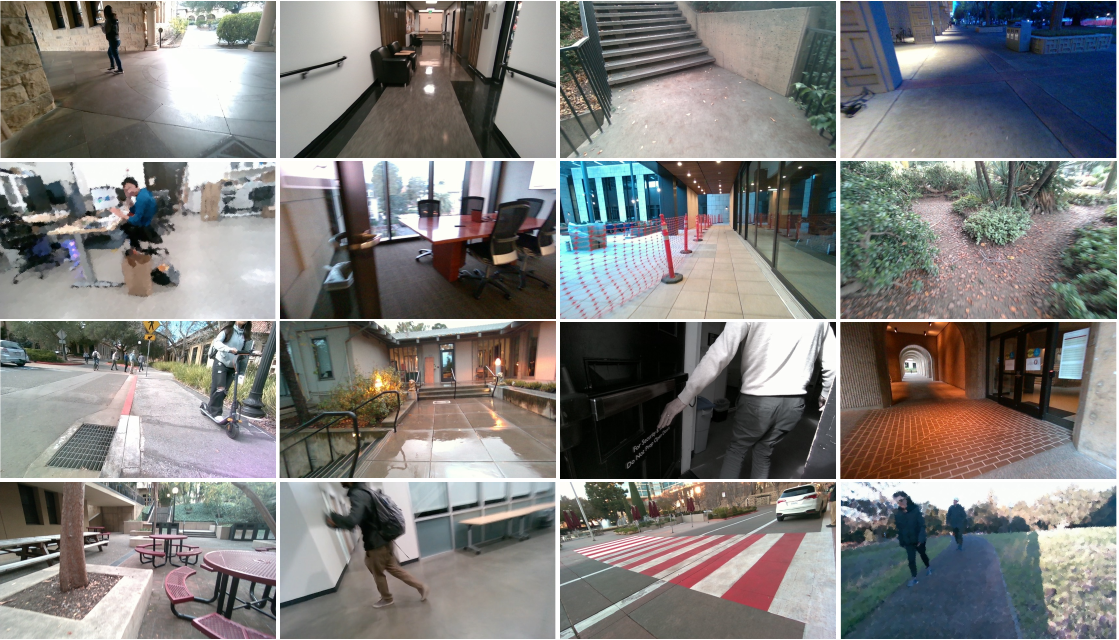}
  \caption{\textbf{Dataset:} The dataset has a mix of weather, road, lighting, and traffic.}
  \label{fig:traj_features}
\vspace{-3mm}
\end{figure}

EgoNav learns a navigation prior entirely from human walking data and deploys it zero-shot on a humanoid robot. The system is designed around four principles:

\begin{itemize}
    \item \textbf{Human-native.} The system is capable of learning entirely from human walking demonstrations, requiring no robot data, simulation, or task-specific engineering. Human data is cheap, scalable, and encodes rich commonsense about navigation.
    \item \textbf{Scene-aware.} The system derives sufficient scene understanding from egocentric observations alone, without requiring pre-built maps, prior environment models, or any external localization.
    \item \textbf{Distributional.} The system models the full distribution of plausible future paths, providing downstream components with diverse candidates for informed selection, rather than committing to a single trajectory estimate.
    \item \textbf{Robot-ready.} The learned prior is deployable on physical robots, not merely performant in offline evaluation. This requires addressing real-time inference speed, system latency compensation, closed-loop mode consistency, and robust obstacle avoidance.
\end{itemize}

Formally, let a trajectory $\tau$ be a sequence of 6D poses in the 3D world. At time $t$, we model the multi-modal distribution of future trajectories conditioned on past trajectory, a visual memory encoding $S$, and egocentric video features $F$:
\begin{equation}
    \hat{\tau}_{t:t+T} \sim p_{\theta}(\cdot | \tau_{1:t},S, F)
    \label{eqn:prob_statement}
\end{equation}
Each sample from $p_{\theta}$ corresponds to a distinct plausible future path. The following sections describe the design decisions we make to realize these goals (\cref{fig:sys_overview}).

\subsection{Data Collection}
\label{sec:dataset}

\textbf{Collection Setup.}
A single person wearing body-mounted cameras can gather hours of navigation data simply by walking through diverse environments, with no robot hardware or teleoperation required. Our dataset, collected under institutional IRB approval(IRB-60675), consists of 44 sequences ($\sim$7 minutes, 600 meters each) recorded at 20\,Hz using an Intel RealSense T265 for SLAM-based 6-DoF localization and a RealSense D455 stereo camera for aligned RGBD (\cref{fig:traj_features}). The full dataset totals 300 minutes and over 25\,km of walking. Despite its modest size, the dataset deliberately maximizes environmental diversity: sequences span indoor lab space, elevator, offices, stairwells, and outdoor crosswalks, off-road and garden paths; surface types include tile, wood, concrete, brick, metal grates, and rough terrain; lighting ranges from bright daylight to nighttime; weather conditions include sun and rain; and traffic varies from empty paths to crowded pedestrian areas with scooters and construction barriers. This breadth of coverage, combined with data augmentation (320k training samples), enables the model to generalize well beyond the training environments. Compared to existing egocentric datasets such as TISS \cite{qiu_egocentric_2022} and Aria Everyday \cite{lv_aria_2024}, ours provides dense ground-truth depth at higher frequency.

\textbf{Preprocessing.}
Color frames are semantically labeled by DINOv2 \cite{oquab_dinov2_2024} with a Mask2Former head into 8 classes (ground, stair, door, wall, obstacle, movable, rough ground, unlabeled), and depth frames are preprocessed with a Canny edge filter to remove stereo artifacts along object boundaries. All past poses are transformed to an egocentric coordinate frame. We extract overlapping sub-trajectories for data augmentation, yielding over 320,000 training samples.

\subsection{Scene Representation}
\label{sec:scene}

\begin{figure}[tb]
  \centering
  \includegraphics[width=\columnwidth]{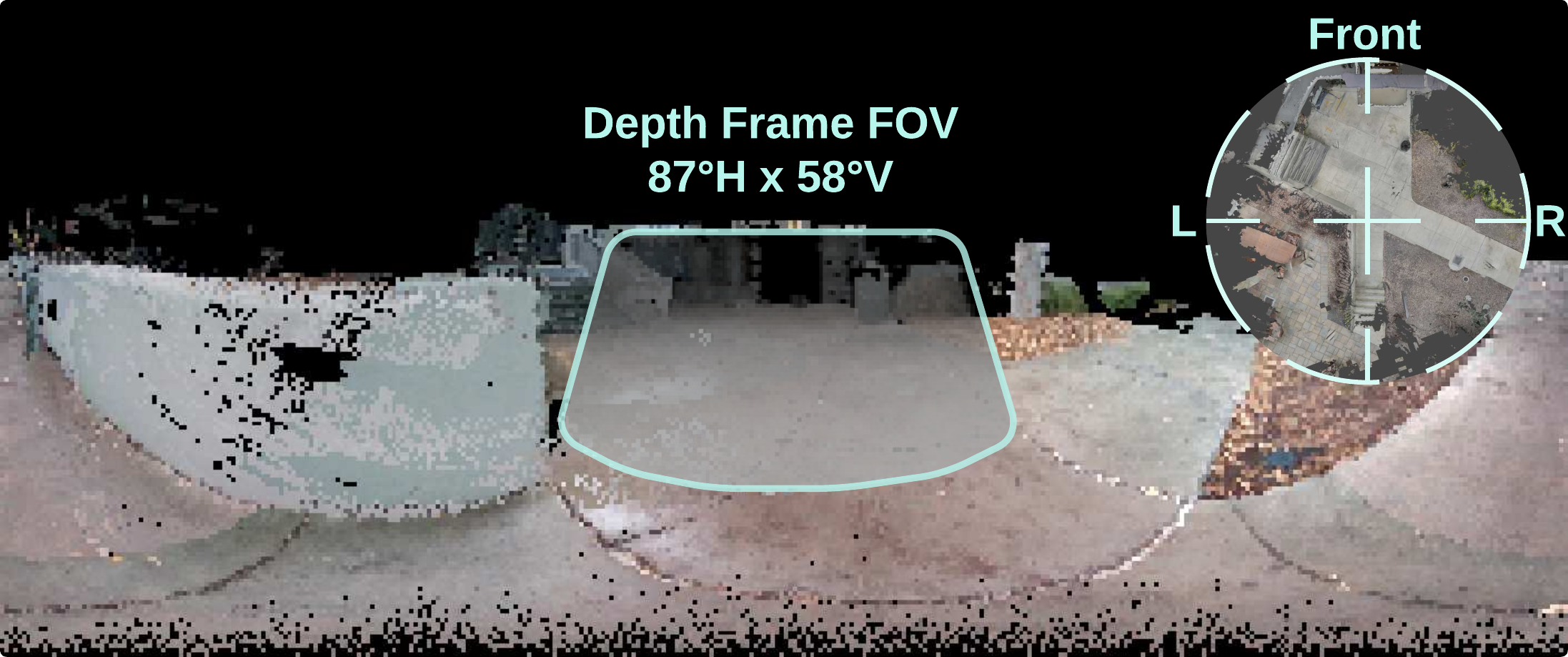}
  \caption{\textbf{Comparing depth frame with visual memory:} A raw depth frame has only $\sim$90\textdegree{} of FOV and misses important scene information. The depth frame sees only the open space ahead and does not capture the stairs, the right turn path, or the wall to the left. Black regions are areas not yet observed.}
  \label{fig:semantic}
\vspace{-3mm}
\end{figure}

\begin{figure}[tb]
  \centering
  \includegraphics[width=\columnwidth]{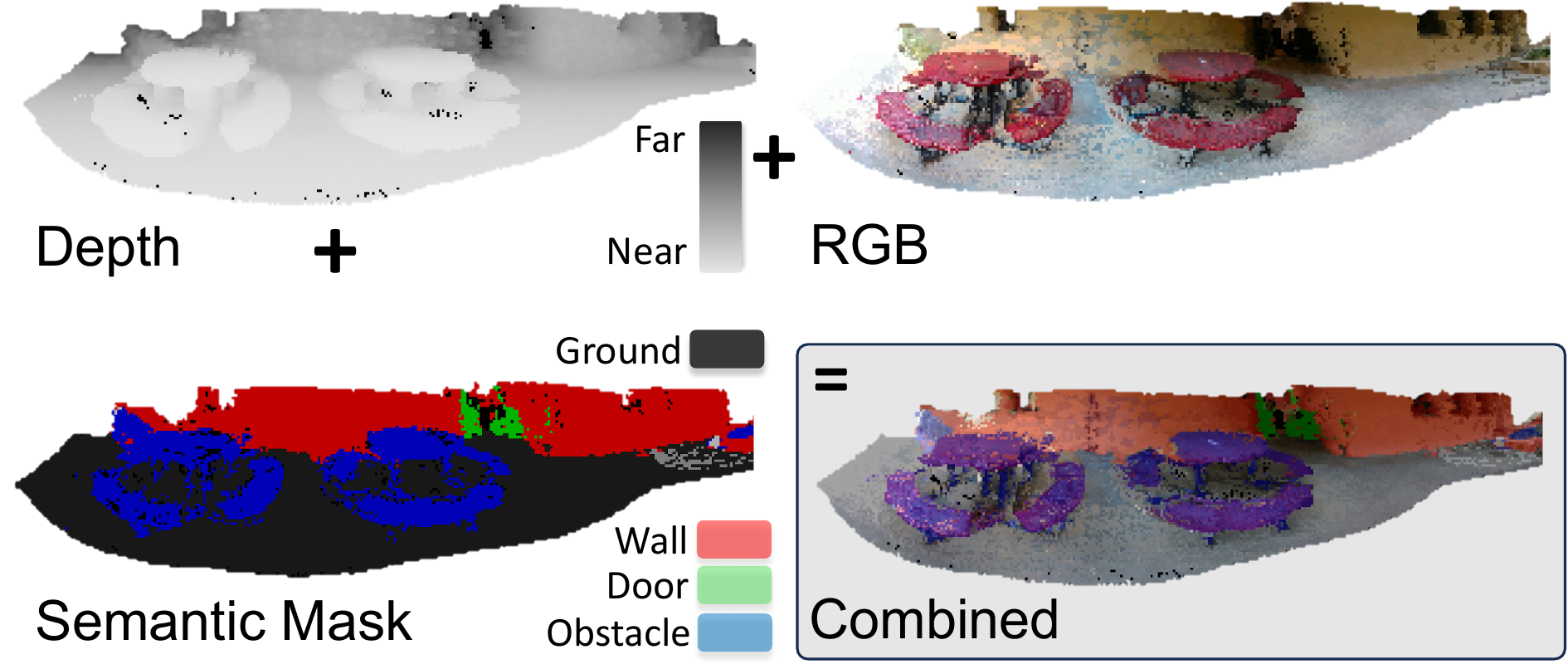}
  \caption{\textbf{Channels in Visual Memory:} The visual memory integrates past frames into a single panorama. It consists of a depth, color, and intensity-encoded 8-class semantic channel. 4 out of 8 channels are shown in the figure.}
  \label{fig:panorama}
\vspace{-3mm}
\end{figure}

\textbf{Visual Memory (VM).}
\label{sec:VM}
A single stereo frame covers only $\sim$90\textdegree{} and misses critical context such as side paths, nearby obstacles, and the space behind the wearer (\cref{fig:semantic}). The visual memory addresses this by accumulating a rolling buffer of RGBD frames into a single 360\textdegree{} egocentric panorama of size $180{\times}360{\times}5$ (R, G, B, depth, semantic; 1 pixel per degree) via 3D reprojection (\cref{fig:panorama}). The VM is encoded by a pretrained spatial VAE (trained with InfoLoss \cite{zhao_infovae_2018}, L1 loss on RGBD channels, and cross-entropy loss on semantic channels) into an $8{\times}20{\times}8$ latent feature map that preserves the geometric layout of the panorama. This spatial encoding is then compressed to a 64-dimensional embedding by a lightweight adapter shared with the DINOv3 branch (\cref{sec:dino}), which aggregates spatial features via learned attention. The VAE remains frozen during diffusion training. VM construction runs on the Jetson Orin NX CPU within 30\,ms. Removing all visual input (VM and video features) produces the worst collision avoidance in our offline evaluation (\cref{sec:evaluation}), confirming the importance of scene context.

\textbf{Egocentric Video Features.}
\label{sec:dino}
While the VM provides a rich spatial summary, it has three blind spots: transparent surfaces such as glass walls are invisible to stereo depth, sensor inaccuracies produce artifacts along object boundaries, and dynamic obstacles that enter or leave the scene between buffer updates are not reliably captured. To complement the VM with appearance-level cues, we extract patch-level features from the most recent egocentric RGB frame using a frozen DINOv3 ViT-S/16 backbone \cite{simeoni_dinov3_2025}, yielding a $15{\times}27$ spatial grid of 384-dimensional patch tokens. A lightweight adapter aggregates these via learned attention: first along the horizontal axis (left--right context), then along the vertical axis (near--far context) with an intermediate 1D convolution, producing a 64-dimensional embedding. At deployment, the smallest ViT-S/16 runs $\sim$6\,ms per frame, adding minimal latency. The impact of these features is validated in real-world deployment (\cref{sec:real_eval}), where they are critical for avoiding obstacles invisible to depth alone.

\subsection{Trajectory Diffusion Model}
\label{sec:diffusion}

\textbf{Architecture.}
The prediction module is a non-autoregressive UNet diffusion model that predicts all 100 future steps (5 seconds at 20\,Hz) simultaneously, ensuring temporal coherence across the full trajectory. Multi-head self-attention (MHSA) layers between downsample and upsample blocks relate conditioning embeddings to different parts of the trajectory. The past trajectory is represented as 100 steps of $x$-$y$-$z$ position and ortho6d \cite{zhou_continuity_2020} orientation (9 channels) at 20\,Hz. All three conditions: VM encoding, video features, and past trajectory, are fused by an adapter module and injected into each UNet block via classifier-free guidance, dropping each condition 10\% of the time during training.


\subsection{Deployment on Unitree G1}
\label{sec:deployment}

\textbf{Hardware.}
The system is deployed on a Unitree G1 humanoid. We mount the data collection cameras directly on the chest of the robot. VM construction, VAE encoding, and trajectory controller run locally on the onboard Jetson Orin NX. Due to constrained compute on the edge device, the diffusion model and semantic segmentation run on a Jetson Thor connected via low-latency local network, which can be carried on the back of the robot for long-range trials. The complete prediction loop runs in real time at $\sim$2\,Hz.

\textbf{Hybrid Generation.}
To achieve real-time inference, we introduce a hybrid sampling scheme that initiates with DDIM steps to quickly approximate the trajectory distribution, followed by DDPM refinement steps to recover fine-grained details. This retains the multi-modal structure of full DDPM while being 100$\times$ faster. The model is trained with a linear noise schedule; at inference, DDIM uses evenly spaced skip steps to rapidly reach a low-noise regime, then DDPM walks consecutive final steps to recover fine details. All three choices: linear schedule, skip-step DDIM, and consecutive-step DDPM finishing, are necessary: replacing the linear schedule (e.g.\ with cosine) breaks the skip-step assumption, and omitting either phase degrades quality significantly. We evaluate step combinations in \cref{sec:evaluation}; the optimal configuration of 5 DDIM + 5 DDPM steps achieves near-full-DDPM quality at real-time rates. On Jetson Thor, the model generates 110 trajectories per second; with a batch size of 64, inference runs at approximately 1.7\,Hz, sufficient for closed-loop control.

\textbf{Receding Horizon Controller.}
\label{sec:rhc}
The navigation prior outputs 6-DoF waypoints in a local egocentric frame. Because this representation specifies \emph{where} to walk without dictating \emph{how}, it serves as an embodiment-agnostic interface: the same model trained on human data transfers to the G1 without finetuning. At each prediction cycle, the diffusion model generates 64 candidate trajectories spanning a 5-second horizon. Trajectories that collide with obstacles in the VM point cloud are filtered out via KD-tree queries. The remaining candidates are clustered via K-Means ($k{=}3$) based on their positions at a 2-second horizon. Each cluster receives two scores: a popularity score (fraction of trajectories in the cluster) and a momentum penalty (distance from the cluster center to the previously selected intention point). The combined score favors large, consistent clusters, preventing erratic switching between modes across cycles. The medoid of the highest-scoring cluster is selected for execution. The controller dynamically estimates the current system latency, discards the corresponding initial segment of the trajectory, and executes the plan with smooth blending against the previous trajectory until the next prediction batch arrives.

\section{Evaluation}
\label{sec:evaluation}

We evaluate EgoNav to answer three questions:
\begin{itemize}
    \item \textbf{Q1}: Does each design component contribute to prediction quality? (\cref{sec:offline_eval})
    \item \textbf{Q2}: Does the multi-modal approach outperform unimodal and non-panoramic baselines? (\cref{sec:offline_eval})
    \item \textbf{Q3}: Can the learned prior, trained entirely on human data, deploy on a real humanoid in unseen environments with zero robot data? (\cref{sec:real_eval})
\end{itemize}

    \subsection{Experimental Setup}
    \label{sec:exp_setup}

    \begin{figure}[tb]
      \centering
      \includegraphics[width=\linewidth]{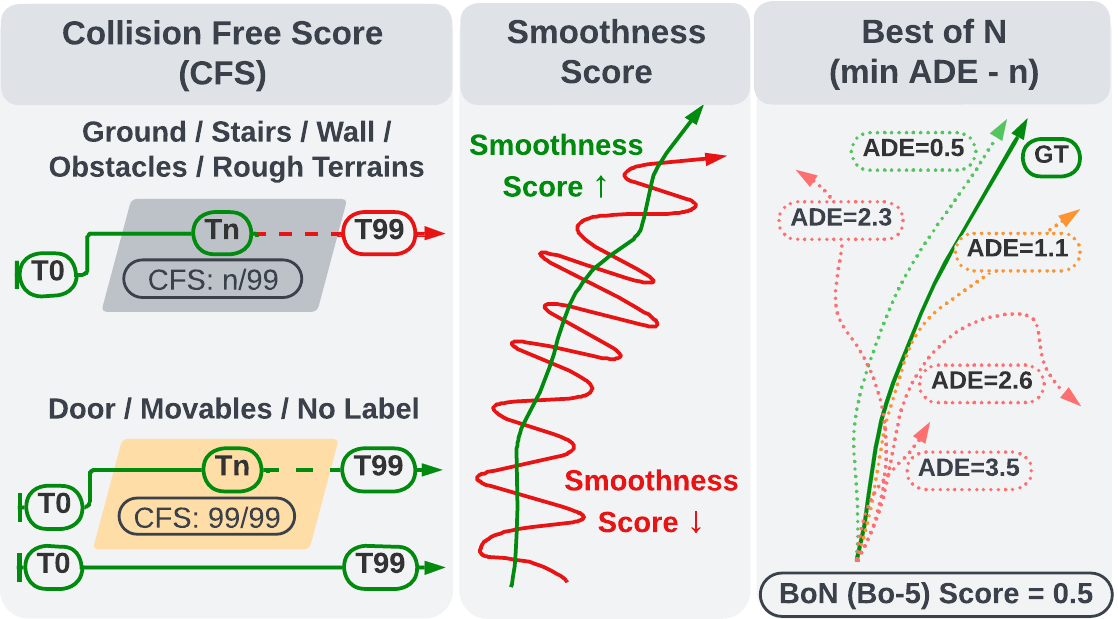}
      \caption{\textbf{Metrics overview:} Collision-free score (CFS) uses selected semantics. All subsequent trajectories will be marked as collided; higher CFS and Smoothness are better, lower Best of N is better.}
      \label{fig:overview_metrics}
      \vspace{-2mm}
    \end{figure}

    \textbf{Metrics.}
    We evaluate predictions using 3 metrics (\cref{fig:overview_metrics}). 
    
    \emph{Collision-free score} measures trajectory feasibility against a point cloud re-projected from the VM: at each time step, we query the 20 nearest points via a KD-tree; a collision is detected if more than 10 points fall within a 16\,cm radius, and all subsequent steps are marked as collided. The score counts consecutive collision-free steps against static obstacles (ground, stairs, walls, rough terrain), excluding doors and movable objects. 
    
    \emph{Smoothness} is the reciprocal of the mean absolute error in speed and acceleration relative to ground truth:
    \begin{align}
        Smoothness = 1/\frac{\sum^{n}_{i=1}(|V_i - \hat{V}_i|+|a_i - \hat{a}_i|)}{n}
    \end{align}
    \emph{Best of N} (minADE-K) selects the closest of K predicted trajectories to ground truth, measuring multi-modal coverage, while \emph{Best of 1} (ADE) evaluates single-prediction accuracy for comparison with unimodal baselines.

    \textbf{Dataset and Training.}
    We report the full system trained on the complete dataset at the top of \cref{tab:ablations}. The w/o DINOv3 variant uses the same checkpoint with video features zeroed out at inference (enabled by classifier-free guidance). The diffusion model has 46M parameters. Component ablations that require retraining (architecture or input changes) and baselines are trained on the pilot subset for faster iteration, sharing the same pretrained VAE encoder for fair comparison. The ground truth (GT) collision score (97.6) is not perfect due to SLAM drift and depth artifacts.

    \subsection{Offline Evaluation}
    \label{sec:offline_eval}

    Ablation and baseline results are summarized in \cref{tab:ablations}.

    \begin{table}[tb]
      \caption{Offline evaluation across ablations and baselines. \textbf{Best}, \underline{\textit{2nd}}.}
      \label{tab:ablations}
      \centering
      \setlength{\tabcolsep}{2pt}
      \resizebox{\columnwidth}{!}{%
        \begin{tabular}{@{}l|cccc@{}}
          \toprule
          Full System (Full Data) & Collision $\uparrow$& Smoothness $\uparrow$& Best of 1 $\downarrow$& Best of 15 $\downarrow$\\
          \midrule
          GT & 97.6 & $\infty$ & 0.0 & 0.0 \\
          Ours     & \textbf{91.4} & \textbf{4.82}  & \textbf{0.76}  & \textbf{0.39} \\
          w/o DINOv3 features     & \underline{\textit{90.6}} & \underline{\textit{4.76}}  & \underline{\textit{0.81}}  & \underline{\textit{0.41}} \\
          Ours - Pilot      & 89.2  & 2.04 & 0.87  & 0.47\\
          \midrule
          Component Ablations (Pilot Data) & & & & \\
          \midrule
          w/o attention     & 86.6 & 2.78 & 1.00 & 0.49 \\
          w/o semantic    & 84.1 & 4.17 & 0.91 & 0.53\\
          w/o VM or DINO (Traj only)     & 82.5 & 2.04 & 1.19 & 0.48 \\
          \midrule
          Baselines (Pilot Data) & & & & \\
          \midrule
          Ours - Pilot      & \textbf{89.2} & 2.04 & \textbf{0.87} & \textbf{0.47}\\
          VAE-LSTM        & \underline{\textit{84.5}} & \textbf{9.09} & \underline{\textit{1.01}} & N/A \\
          CXA Transformer & 80.3 & \underline{\textit{3.70}} & 1.25 & N/A \\
        \bottomrule
        \end{tabular}
      }
    \vspace{-3mm}
    \end{table}

    \textbf{Scene Understanding.}
    Complete removal of visual input yields the worst-performing model (collision 82.5), confirming the importance of scene context; without the VM, generated samples largely memorize training trajectories and produce incorrect mode distributions. Removing the semantic channel causes a significant drop in collision avoidance ($-5.1$): without semantics, the model cannot differentiate between doors and walls, and conflicting training signals cause it to ignore geometric constraints in the VM.

    \textbf{Model Architecture.}
    Adding multi-head self-attention (MHSA) layers between down and up blocks improves collision avoidance by $+2.6$ with negligible impact on inference speed, confirming that attention helps relate VM context to trajectory samples.

    \textbf{DINOv3 Video Features.}
    Zeroing out DINOv3 features at inference causes a modest offline drop (91.4 $\rightarrow$ 90.6 collision). This is expected: the offline collision metric relies on the same depth-based point cloud which does not see glass or dynamic obstacles where the DINOv3 variant excels. The critical role of DINOv3 emerges in real-world deployment, as analyzed in \cref{sec:real_eval}.

    \textbf{Data Scaling.}
    Training on the full dataset improves all metrics over the pilot subset (collision 89.2 $\rightarrow$ 91.4, Best of 15 0.47 $\rightarrow$ 0.39), confirming that our navigation prior is scalable and can benefit from more diverse walking data.

\begin{figure}[tb]
      \centering
      \includegraphics[width=\columnwidth]{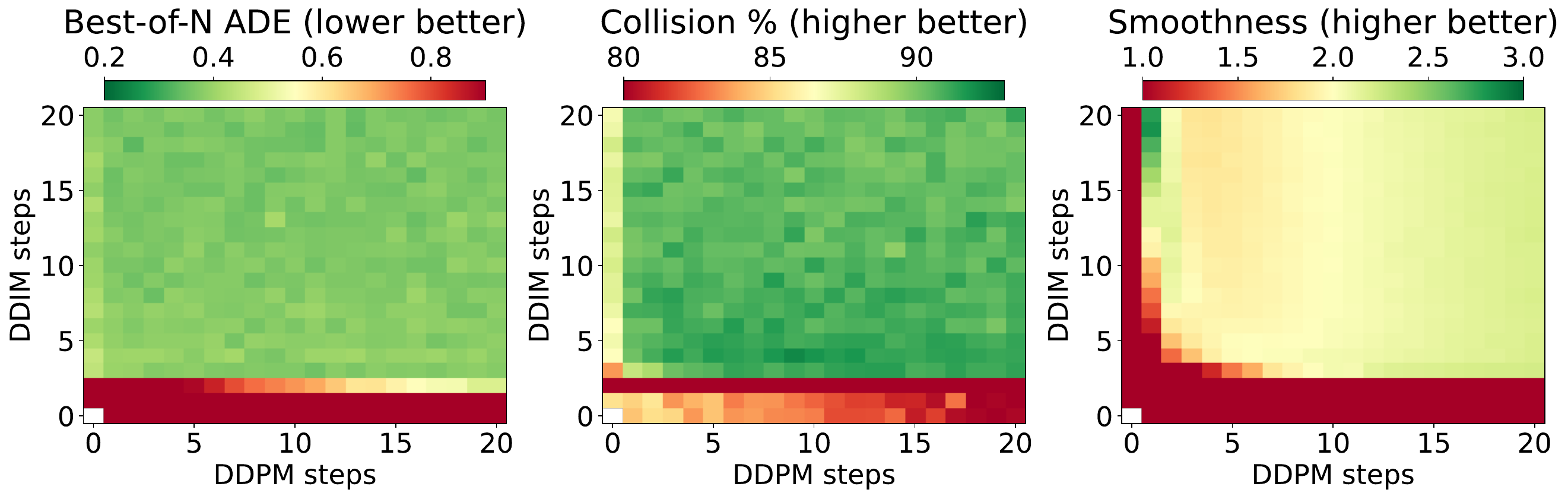}
      \caption{\textbf{Hybrid generation step search:} BoN, Collision, Smoothness score for different DDIM/DDPM step combinations. The optimal configuration is 5 DDIM + 5 DDPM. All sweeps are performed on the same checkpoint. Combination 0+0 is pure noise therefore omitted.}
      \label{fig:hybrid_steps}
    \vspace{-3mm}
    \end{figure}

    \textbf{Hybrid Generation.}
    We systematically evaluate different combinations of DDIM and DDPM steps (\cref{fig:hybrid_steps}). The optimal configuration of 5 DDIM + 5 DDPM steps achieves near-full-DDPM quality with 100$\times$ fewer steps. Fewer than 7 total steps degrade smoothness significantly. Pure DDIM with the same total steps performs notably worse in both collision and smoothness, confirming the value of the DDPM refinement.

    \noindent\textbf{Comparison with Baselines}

    \textbf{VAE-LSTM \cite{wang_trajectory_2023}} auto-regressively predicts future states, producing deterministic, unimodal outputs. Its step-by-step generation excels at smooth motion ($+7.05$), and it demonstrates surprisingly robust obstacle avoidance despite fewer than 1M parameters, likely due to its panorama depth inputs providing broad environmental context similar to our VM. However, it cannot represent multi-modal distributions and requires one forward pass per predicted step, making it the slowest of the three.
    
    \textbf{CXA-Transformer \cite{qiu_egocentric_2022}} employs cascaded cross-attention to fuse pedestrian poses, semantic segmentation, and past trajectories. Its collision score is significantly lower ($-8.9$) than ours, and its ADE is worse ($+0.38$), though it achieves higher smoothness ($+1.66$). The original method uses 2\,Hz RGBD directly without a panoramic representation, which limits situational awareness; without the structured scene context that a VM provides, the model likely requires significantly more data to learn robust obstacle avoidance. Like LSTM, it produces deterministic predictions.
    Overall, our method achieves the best collision avoidance and mode coverage among all compared methods.

    \subsection{Real-world Deployment}
    \label{sec:real_eval}

    To validate EgoNav beyond offline metrics, we deploy the full system zero-shot on a Unitree G1 humanoid robot in unseen environments. We do not compare against baselines, as they either lack a real-time system (LookOut, CXA, VAE-LSTM) or require robot-collected data (NoMaD, HEAD). We evaluate across four categories (\cref{tab:real_world}): static indoor scenes (kitchen, lab), corridors, areas with glass walls, and dynamic scenes (pedestrians, moving objects). Over 37.5 minutes of autonomous operation covering 1,137\,m, the system achieves 96--99\% autonomous time, with corridors reaching 99\% and the more challenging glass and dynamic scenes slightly lower.

\begin{figure*}[tb]
      \centering
      \includegraphics[width=\textwidth]{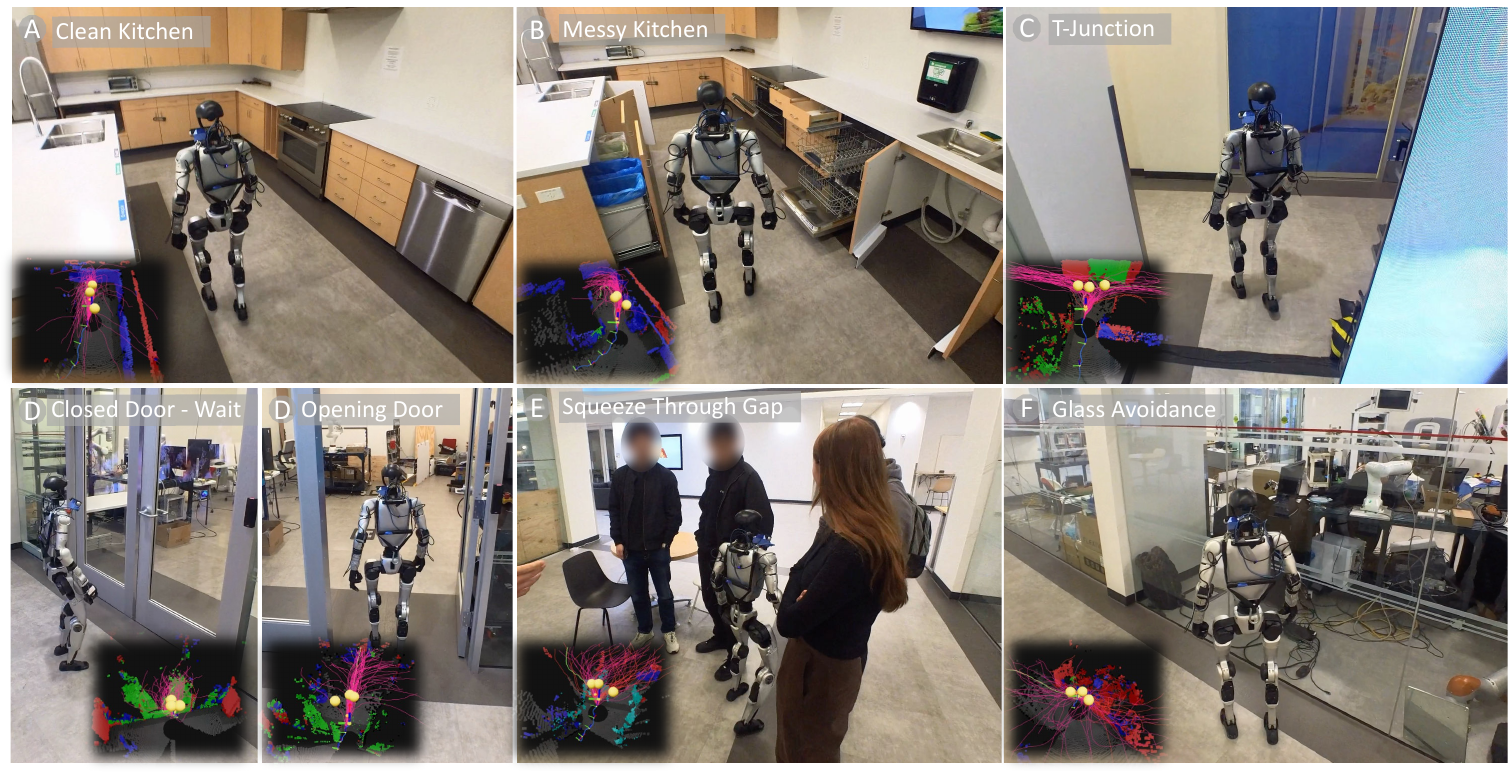}
      \vspace{-7mm}
      \caption{\textbf{Real-world deployment on Unitree G1.} The humanoid navigates diverse unseen environments using EgoNav with zero robot training data. Insets show predicted trajectory distributions overlaid on the semantic point cloud. (A,B)~Static indoor scenes with different clutter levels. (C)~T-junction with multi-modal trajectory predictions. (D)~The robot waits at a closed door and proceeds when it opens, demonstrating semantic understanding. (E)~Navigating through a gap between pedestrians. (F)~Avoiding a glass wall invisible to depth sensors, enabled by DINOv3 video features.}
      \label{fig:g1_deployment}
    \vspace{-3mm}
    \end{figure*}

    During static scene evaluation, we randomly rearrange objects to test generalization. The system handles the kitchen even with all cabinets pulled out (\cref{fig:g1_deployment}B). The supplementary video shows robustness in more scenes where furniture is moved to random locations. At a T-junction (\cref{fig:g1_deployment}C), EgoNav predicts a bimodal distribution and the momentum penalty commits to turning left. The system also exhibits surprising dynamic scene behaviors. In \cref{fig:g1_deployment}D, it waits at a closed door until it opens, then proceeds when the path is clear. In a crowd gathering scene (\cref{fig:g1_deployment}E), it finds narrow gaps between pedestrians and navigates through.

    DINOv3 features prove critical for glass environments: in \cref{fig:g1_deployment}F the system turns away from a glass wall invisible to depth sensors. Without DINOv3, the intervention rate in glass environments nearly triples (0.77 $\rightarrow$ 2.2/min, \cref{tab:real_world}).

    \begin{table}[tb]
      \caption{Real-world deployment statistics on the Unitree G1.}
      \label{tab:real_world}
      \centering
      \small
        \begin{tabular}{@{}lcccc@{}}
          \toprule
           & Static & Corridor & Glass & Dynamic \\
          \midrule
          Duration (min)    & 16 & 7.5 & 6.5 (5) & 7.5 \\
          Distance (m)      & 471 & 240 & 225 (120) & 201 \\
          Interventions     & 9 & 1 & 5 (11) & 6 \\
          Interventions/min & 0.56 & 0.13 & 0.77 (2.2) & 0.80 \\
          Autonomous time\%     & 97.2 & 99.3 & 96.2 (89.0) & 96.0 \\
          \bottomrule
        \end{tabular}\\[4pt]
        \footnotesize Glass column: values in parentheses are w/o DINOv3.
    \vspace{-3mm}
    \end{table}

    \textbf{Failure Mode Analysis.}
    We also deployed ablation versions to isolate each component. Without DINOv3, glass walls invisible to depth cause the intervention rate to nearly triple (0.77 $\rightarrow$ 2.2/min, \cref{tab:real_world}), confirming that the modest offline delta masks a critical real-world contribution. Without semantic labels, geometrically similar doors and walls become indistinguishable, and the model frequently predicts paths through solid walls. Without the momentum penalty, trajectory clusters alternate between modes across cycles, producing oscillatory behavior at decision points.

\section{Conclusion}
\label{sec:conclusion}
We presented EgoNav, a system that learns humanoid navigation entirely from human walking data, requiring no robot data or finetuning. A conditional diffusion model, conditioned on a 360\textdegree{} visual memory and DINOv3 video features, internalizes commonsense navigation behavior from egocentric observations. A hybrid sampling scheme enables real-time inference. Through offline evaluations and zero-shot deployment on a Unitree G1 in unseen environments, we show that behaviors such as waiting at doors, navigating around pedestrians, and avoiding transparent obstacles emerge from the learned prior without explicit programming.

Future work includes goal-conditioned trajectory selection for directed navigation, monocular depth estimation to relax the stereo camera requirement, and scaling to more diverse environments. More broadly, a navigation prior can serve as the missing middle layer between high-level task planning and low-level locomotion: the predicted paths tell a controller where to go, while the semantic visual memory along those paths informs which locomotion skill to use.


\bibliographystyle{IEEEtran}
\bibliography{KenBib,egonav_resubmit}

\clearpage
\newpage
\appendices

\end{document}